\documentclass[journal,10pt,letterpaper]{IEEEtran}

\usepackage[T1]{fontenc}
\usepackage{amsmath,amssymb}
\usepackage{graphicx}
\usepackage{xcolor}
\usepackage{booktabs}
\usepackage{cite}
\usepackage{algorithm}
\usepackage{algorithmic}
\usepackage[hidelinks]{hyperref}

\title{TRACE: Two-Stage Detector-Response Estimation With Angular Cosine Expansion for Ring Artifact Correction in Photon-Counting CT}
\author{
Jigang Duan, Heran Wang, Ligen Shi, Zheng Sun, Ping Yang, and Xing Zhao%
\thanks{Jigang Duan, Heran Wang, Ligen Shi, Zheng Sun,
Ping Yang, and Xing Zhao are with the School of Mathematical
Sciences, Capital Normal University, Beijing 100048, China.}%
\thanks{Ligen Shi is with the College of Computer Science (College of Software), Inner Mongolia University,
Hohhot 010021, China, and also with the Research Center for Spatiotemporal Intelligence, Inner Mongolia University, Hohhot 010021, China}%
\thanks{Corresponding authors: Ping Yang and Xing Zhao
(e-mail: \texttt{yang\_ping0603@163.com};
\texttt{zhaoxing\_1999@126.com}).}%
}
\begin{document}

\maketitle

\begin{abstract}
Detector response nonuniformity introduces systematic projection errors and ring artifacts in photon-counting detector computed tomography (PCD-CT). In measured PCD-CT data, residual stripe amplitudes vary slowly with projection angle, which fixed-bias models cannot adequately capture. We propose TRACE, a two-stage unsupervised sinogram decomposition method for estimating and correcting these response-related errors. TRACE represents stripes as a fixed bias plus low-order discrete cosine transform (DCT) components, using a small number of coefficients to describe angular variations at each detector element. A learnable analysis--synthesis architecture represents the ideal projections, while two-stage optimization separates them from fixed and then dynamic stripes. An angular-gradient soft orthogonality constraint suppresses correlated variations within the shared DCT gradient subspace, reducing the leakage of object structures into the artifact estimate. All parameters are optimized directly on the measured sinogram without paired training data. Experiments on measured QRM mouse phantom and porcine trotter data show that TRACE suppresses ring artifacts and improves image uniformity while preserving edge sharpness, soft-tissue texture, and trabecular detail.
\end{abstract}

\begin{IEEEkeywords}
Photon-counting CT, detector response nonuniformity, ring artifact correction, sinogram decomposition, discrete cosine transform, unsupervised learning
\end{IEEEkeywords}

\section{Introduction}

Response nonuniformity among detector elements introduces systematic errors into computed tomography (CT) measurements. These errors appear as stripes extending along the angular direction in sinograms and form rings around the rotation center after reconstruction~\cite{sijbers2004reduction,munch2009stripe}. Unlike conventional energy-integrating detector CT, photon-counting detector CT (PCD-CT) registers individual X-ray-induced electrical pulses and discriminates their amplitudes using energy thresholds. Differences in counting response between detector elements can depend on energy threshold and incident flux, so fixed calibration parameters may not fully compensate for response nonuniformity throughout a scan. Residual response errors can produce stripes that vary with energy window and projection angle~\cite{chen2026physics,lee2026calibration,hsieh2026autocalibration}. These artifacts impair image uniformity, distort reconstructed attenuation values, and obscure low-contrast structures. Correcting residual response errors while preserving object information is therefore important for reliable PCD-CT measurements.

\subsection{Existing Methods for Ring Artifact Correction}

Existing methods include detector calibration, sinogram-domain preprocessing, CT image postprocessing, dual-domain iterative methods, and data-driven methods.

Detector calibration compensates for response differences between detector elements through dynamic flat-field correction based on eigen flat fields~\cite{vannieuwenhove2015dynamic} or nonlinear response fitting based on mean projections~\cite{guo2022nonlinear}. For PCD-CT, phantom-based calibration corrects response differences and count-rate nonlinearity~\cite{lee2026calibration}, while physical models estimate energy threshold deviations~\cite{chen2026threshold}. Redundant ray sampling also enables gain self-calibration from scan data~\cite{hsieh2026autocalibration}. However, fixed parameters may fail to track changing system and scan conditions, whereas dynamic calibration often requires additional data, redundant sampling, or accurate registration.

Sinogram-domain preprocessing suppresses stripes before reconstruction through wavelet--Fourier filtering~\cite{munch2009stripe}, extensions using projection grouping and weighted filtering~\cite{guo2015ct}, or response equalization, detection, and interpolation~\cite{vo2018superior}. For PCD-CT, smoothed mean projections can provide detector response correction coefficients~\cite{an2020ring}. Other approaches use multiscale BM3D for correlated stripe noise~\cite{makinen2021multiscale}, low-rank Tucker decomposition with spatial--sequential total variation~\cite{li2022lowrank}, adaptive normalization~\cite{kazimirov2024adaptive}, or adaptive frequency-domain patch filtering combined with spatial-domain stripe filtering~\cite{liu2025adaptive}. However, directional or frequency overlap between stripes and object projections can lead to residual artifacts or loss of useful projection information.

CT image postprocessing typically converts rings into approximate stripes in polar coordinates. Representative methods use sliding-window artifact estimation~\cite{sijbers2004reduction}, unidirectional variational decomposition with sparsity constraints~\cite{yan2016variation}, or relative total variation (RTV) templates for iterative stripe extraction~\cite{liang2017iterative}. Structure preservation is addressed through sparse-domain regularized decomposition with guided filtering~\cite{li2021sdrsd}, structure-aware guided filtering~\cite{zhao2023structureaware}, or directional gradient-domain optimization~\cite{wang2024analytical}. Superpixel segmentation with adaptive RTV further treats strong and weak rings separately~\cite{li2025superpixel}. These methods require no raw projections, but polar-coordinate interpolation may blur details, and circular object structures may be suppressed. They also lack direct constraints from measured projections.

Dual-domain iterative methods jointly optimize image reconstruction and projection correction. Early studies introduced explicit ring variables into compressed sensing reconstruction~\cite{paleo2015ring} or combined image-domain ring total variation with detector correction vectors~\cite{salehjahromi2019ring}. Extensions incorporate unidirectional total variation and group sparsity into spectral PCD-CT material decomposition~\cite{sun2025spectral}, combine stripe and image sparsity priors~\cite{lu2025dual}, or couple polynomial detector response models with polar-domain smoothing~\cite{kan2025dudo}. Other formulations combine group sparsity and low-rank priors on three-dimensional projection stripes with image regularization~\cite{qin2026dual}, or integrate generative and ring artifact model priors for sparse-view reconstruction~\cite{li2025modelguided}. Repeated projection, backprojection, and subproblem optimization increase computational cost, while performance depends on the system model and regularization. Separate component priors may also permit object structures to enter artifact estimates.

Data-driven methods learn correction mappings or artifact representations using neural networks. Early approaches fused precorrected sinograms with reconstructed images~\cite{chang2021hybrid} or combined random detector translation with neural artifact suppression and denoising~\cite{liu2023detector}. Sinogram networks estimate stripes by combining wavelet decomposition with residual learning~\cite{fu2023deep} or global--local feature interaction~\cite{su2025globallocal}. Image-domain methods include a polar-domain Transformer with a vertical-gradient loss~\cite{sha2024transformer}, Cartesian--polar Mamba networks~\cite{zhang2025gdp}, and dual-branch networks exploiting adjacent slices and enhancing central image regions~\cite{zhang2025ice}. Multistage networks also process projections, sinograms, and reconstructed images sequentially~\cite{shi2025multistage}. Syn2Real uses realistic image-domain ring synthesis to reduce the synthetic--real gap~\cite{hein2025syn2real}, while conditional flow matching refines reconstructions after random neighborhood sampling decouples projection stripes~\cite{li2026flow}.

Unsupervised alternatives use dual contrastive learning for unpaired polar-domain correction~\cite{wang2023unsupervised} or implicit neural representations for single-sinogram decomposition into ideal projections and stripes~\cite{shi2025inr}. Supervised methods depend on training data size, quality, and representativeness and remain sensitive to synthetic--real distribution differences. Unsupervised methods reduce reliance on paired data but still depend on network representations, losses, and priors.

Overall, two challenges remain difficult to address simultaneously: modeling residual detector response variations along the scan sequence with controlled degrees of freedom, and preventing object structures from entering artifact estimates when stripes and object projections exhibit similar variation patterns.

\subsection{Motivation}

\begin{figure}[t]
    \centering
    \includegraphics[width=\columnwidth]{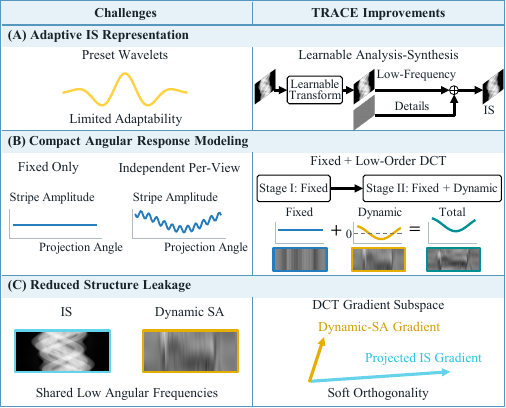}
    \caption{Motivation and key ideas of TRACE.
    (A) A learnable analysis--synthesis representation combines a
    low-frequency backbone with recovered details to estimate the
    ideal sinogram (IS).
    (B) Two-stage optimization extends fixed stripe biases with
    low-order discrete cosine transform (DCT) components to capture
    smooth angular variations.
    (C) Soft orthogonality between the projected angular gradients
    of IS and dynamic stripe artifacts (SA) reduces structure leakage.
    The cyan and yellow vectors represent these gradients, respectively,
    in the subspace spanned by the angular gradients of the low-order
    DCT basis functions.}
    \label{fig:motivation}
\end{figure}

To address these challenges, we propose TRACE, a two-stage unsupervised method that estimates residual detector response errors through sinogram decomposition. Fig.~\ref{fig:motivation} summarizes its three key ideas: an adaptive representation of ideal projections, compact angular response modeling, and reduced structure leakage.

To relate residual response errors to projection stripes, consider an equivalent monoenergetic model for a fixed energy channel. Omitting the angle index and random counting fluctuations, the transmitted count at detector element $d$ is modeled as
\begin{equation}
I_d
=
C_d I_{0,d}
\exp\left(
-\int_{L_d} \mu(x)\,\mathrm{d}l
\right),
\label{eq:motivation_count_model}
\end{equation}
where $I_{0,d}$ is the incident reference count and $C_d>0$ is the effective relative response coefficient after conventional calibration, with an ideal value of 1. The linear attenuation coefficient is denoted by $\mu(x)$, and $L_d$ is the corresponding ray path. This effective model links residual response errors to additive biases after logarithmic conversion:
\begin{equation}
\int_{L_d} \mu(x)\,\mathrm{d}l
=
-\log\left(\frac{I_d}{I_{0,d}}\right)
+
\log\left(C_d\right).
\label{eq:motivation_log_model}
\end{equation}
Thus, the measured logarithmic projection is the sum of the ideal line integral and the stripe component $-\log(C_d)$. Across all views and detector elements, these components form the ideal sinogram (IS) and stripe artifacts (SA), respectively.

Along the detector direction, adjacent object projections are usually locally continuous, giving the IS a predominantly low-frequency structure. Narrow stripes exhibit more rapid variations between detector elements. As illustrated in Fig.~\ref{fig:motivation}(A), predefined wavelet bases can separate frequency components~\cite{munch2009stripe}, but their fixed supports and frequency responses limit adaptation to different object structures and stripe widths. A learnable analysis--synthesis representation can adapt to the measured sinogram while a detail branch recovers structural information.

Along the angular direction, the residual stripes considered here contain fixed biases and slowly varying components. Fig.~\ref{fig:motivation}(B) illustrates the modeling trade-off: fixed biases cannot capture angular variations, whereas independent per-view estimates introduce many degrees of freedom and may absorb object structures. Moreover, IS and dynamic SA can share low angular frequencies, as shown in Fig.~\ref{fig:motivation}(C). Separate smoothness or sparsity constraints therefore do not fully resolve their ambiguity.

Based on these observations, TRACE first estimates fixed stripes and then introduces dynamic components. In Stage I, the learnable analysis--synthesis architecture represents IS, while each detector element's logarithmic response error is approximated by a fixed bias:
\begin{equation}
-\log\left(C_d\right)
\approx
b_d,
\label{eq:motivation_static_model}
\end{equation}
where $b_d$ is the fixed stripe coefficient. Fitting the sum of IS and fixed SA to the measured sinogram limits the initial degrees of freedom of the artifact branch and provides a stable starting point for dynamic estimation.

Following Fig.~\ref{fig:motivation}(B), Stage II extends the response model using low-order discrete cosine transform (DCT) components:
\begin{equation}
-\log\left[C_d(v)\right]
\approx
b_d
+
\sum_{k=1}^{K}
a_{d,k}\phi_k(v),
\label{eq:motivation_dynamic_model}
\end{equation}
where $v$ is the view index, $\phi_k(v)$ is the $k$th non-DC DCT basis function, and $a_{d,k}$ is its coefficient. The number of basis functions, $K$, is much smaller than the number of views. Excluding the DC basis avoids redundancy with the fixed bias, yielding a compact representation of smooth angular response variations.

Finally, Fig.~\ref{fig:motivation}(C) illustrates the angular-gradient soft orthogonality constraint. The angular gradients of IS and dynamic SA are projected onto the subspace spanned by the gradients of the retained DCT basis functions. Penalizing their normalized correlation reduces the leakage of object structures into dynamic SA. Angular curvature regularization controls oscillations, while data consistency and sorting-based detector-direction smoothness stabilize the decomposition.

\subsection{Our Contributions}

The main contributions are as follows:
\begin{itemize}
    \item We develop a two-stage unsupervised sinogram decomposition framework that combines a learnable analysis--synthesis representation with progressive fixed and dynamic stripe estimation. All parameters are fitted to the measured sinogram without paired data or additional flat-field acquisitions.

    \item A fixed--dynamic response-error model uses a truncated non-DC DCT basis to describe smooth angular stripe variations with few coefficients per detector element, extending fixed-bias correction while controlling model complexity.

    \item An angular-gradient soft orthogonality constraint reduces ambiguity between IS and dynamic SA. Together with angular curvature regularization, data consistency, and detector-direction smoothness, it promotes artifact separation while preserving object structures.
\end{itemize}

\section{Method}
\label{sec:method}

\begin{figure*}[t]
    \centering
    \includegraphics[width=\textwidth]{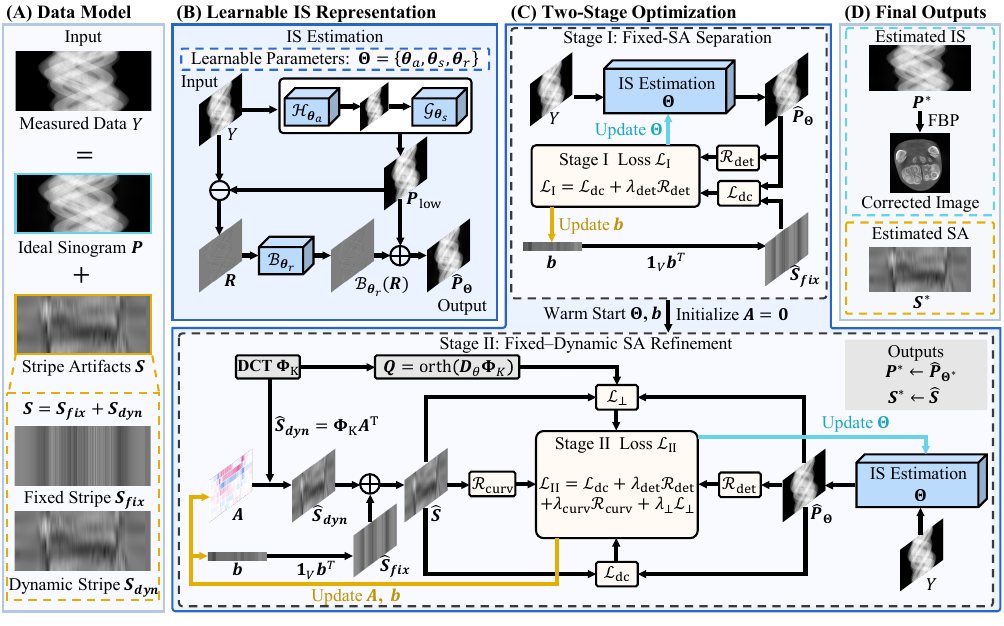}
    \caption{Overview of the proposed two-stage unsupervised sinogram decomposition method.
    (A) IS--SA decomposition of the measured sinogram and the fixed--dynamic stripe model.
    (B) IS representation using a learnable analysis--synthesis architecture and a residual branch.
    (C) Two-stage optimization: fixed stripes are first separated, followed by the introduction of dynamic DCT components and angular constraints to jointly refine IS and SA.
    (D) Final IS and SA estimates and the corrected image reconstructed from IS using FBP.}
    \label{fig:framework}
\end{figure*}

As shown in Fig.~\ref{fig:framework}, TRACE separates IS and SA by exploiting their variation patterns along the detector and angular directions. A learnable IS branch represents object projections, while two-stage optimization progressively estimates fixed and dynamic stripes. All parameters are optimized directly on the measured sinogram.

\subsection{Sinogram Decomposition and Learnable IS Representation}
\label{subsec:is_representation}

As shown in Fig.~\ref{fig:framework}(A), let the measured logarithmic sinogram be
$\boldsymbol{Y}\in\mathbb{R}^{V\times D}$, where $V$ and $D$
denote the numbers of projection views and detector elements, respectively. We use the additive decomposition
\begin{equation}
\boldsymbol{Y}=\boldsymbol{P}+\boldsymbol{S},
\label{eq:sinogram_decomposition}
\end{equation}
where $\boldsymbol{P}$ and $\boldsymbol{S}$ denote IS and SA, respectively.

Fig.~\ref{fig:framework}(B) details the IS branch.
Since IS typically exhibits local continuity and predominantly low-frequency content along the detector direction,
a learnable convolutional analysis--synthesis architecture first extracts its low-frequency backbone:
\begin{equation}
\boldsymbol{P}_{\mathrm{low}}
=
\mathcal{G}_{\boldsymbol{\theta}_{s}}
\left[
\mathcal{H}_{\boldsymbol{\theta}_{a}}(\boldsymbol{Y})
\right],
\label{eq:learnable_analysis_synthesis}
\end{equation}
where $\mathcal{H}_{\boldsymbol{\theta}_{a}}$
comprises analysis convolutions and downsampling, and
$\mathcal{G}_{\boldsymbol{\theta}_{s}}$
comprises upsampling and synthesis convolutions.
Inspired by wavelet analysis and synthesis, this architecture uses learnable convolution kernels to adapt the representation to the measured sinogram.

To recover structural details lost during analysis and synthesis, we compute the residual
\begin{equation}
\boldsymbol{R}
=
\boldsymbol{Y}-\boldsymbol{P}_{\mathrm{low}}.
\label{eq:is_residual}
\end{equation}
As shown in the lower branch of Fig.~\ref{fig:framework}(B),
the residual passes through a learnable vertical band-stop module
$\mathcal{B}_{\boldsymbol{\theta}_{r}}$
to suppress stripes extending along the angular direction.
The result is added to the low-frequency backbone to obtain the IS estimate:
\begin{equation}
\widehat{\boldsymbol{P}}_{\boldsymbol{\Theta}}
=
\boldsymbol{P}_{\mathrm{low}}
+
\mathcal{B}_{\boldsymbol{\theta}_{r}}(\boldsymbol{R}),
\label{eq:is_representation}
\end{equation}
where
$\boldsymbol{\Theta}
=
\{\boldsymbol{\theta}_{a},
\boldsymbol{\theta}_{s},
\boldsymbol{\theta}_{r}\}$
contains all learnable parameters of the IS branch.

\subsection{Fixed--Dynamic Modeling of SA}
\label{subsec:sa_model}

Residual response errors can contain both fixed biases and smooth angular variations.
Following the stripe decomposition in Fig.~\ref{fig:framework}(A),
we represent SA using fixed column biases and low-order DCT components:
\begin{equation}
\widehat{\boldsymbol{S}}(\boldsymbol{b},\boldsymbol{A})
=
\underbrace{
\boldsymbol{1}_{V}\boldsymbol{b}^{T}
}_{\widehat{\boldsymbol{S}}_{\mathrm{fix}}}
+
\underbrace{
\boldsymbol{\Phi}_{K}\boldsymbol{A}^{T}
}_{\widehat{\boldsymbol{S}}_{\mathrm{dyn}}},
\label{eq:fixed_dynamic_sa}
\end{equation}
where $\boldsymbol{1}_{V}$ is an all-ones vector,
$\boldsymbol{b}\in\mathbb{R}^{D}$ contains the fixed biases,
$\boldsymbol{A}\in\mathbb{R}^{D\times K}$ is the dynamic coefficient matrix, and
$\boldsymbol{\Phi}_{K}\in\mathbb{R}^{V\times K}$
contains the first $K$ non-DC DCT basis functions, with entries
\begin{equation}
[\boldsymbol{\Phi}_{K}]_{v,k}
=
\alpha_{k}
\cos\left[
\frac{\pi(2v+1)k}{2V}
\right],
\label{eq:dct_basis}
\end{equation}
where $v=0,\ldots,V-1$, $k=1,\ldots,K$,
$\alpha_{k}$ is a normalization coefficient, and $K\ll V$.
Here, $[\boldsymbol{\Phi}_{K}]_{v,k}=\phi_k(v)$ in~\eqref{eq:motivation_dynamic_model}, and $d=1,\ldots,D$ indexes detector elements.
Excluding the DC basis yields
$\boldsymbol{\Phi}_{K}^{T}\boldsymbol{1}_{V}
=\boldsymbol{0}$,
so the dynamic component has zero angular mean,
avoiding redundancy with the fixed biases.

In Stage II of Fig.~\ref{fig:framework}(C), replicating $\boldsymbol{b}$ across views forms fixed SA, while $\boldsymbol{\Phi}_{K}\boldsymbol{A}^{T}$ generates dynamic SA. Their sum describes the response-related projection bias at each detector element.

\subsection{Regularization and Two-Stage Optimization}
\label{subsec:optimization}

IS--SA separation is guided by data consistency and directional constraints.
First, data consistency penalizes the difference between their sum and the measured sinogram:
\begin{equation}
\mathcal{L}_{\mathrm{dc}}(\boldsymbol{P},\boldsymbol{S})
=
\frac{1}{VD}
\left\|
\boldsymbol{Y}-\boldsymbol{P}-\boldsymbol{S}
\right\|_{1}.
\label{eq:data_consistency}
\end{equation}
Here, $\|\cdot\|_1$ denotes the entrywise $L_1$ norm.

To reduce the influence of angle-dependent shifts in projection structures when comparing adjacent detector elements,
we first sort the projection values in each IS column along the angular direction:
\begin{equation}
\widetilde{\boldsymbol{P}}
=
\operatorname{sort}_{v}(\boldsymbol{P}).
\label{eq:sorted_is}
\end{equation}
Differences between adjacent sorted columns are then penalized
using a sorting-based detector-direction smoothness constraint:
\begin{equation}
\mathcal{R}_{\mathrm{det}}(\boldsymbol{P})
=
\frac{1}{V(D-1)}
\left\|
\widetilde{\boldsymbol{P}}\boldsymbol{D}_{d}^{T}
\right\|_{F}^{2},
\label{eq:detector_smoothness}
\end{equation}
where
$\boldsymbol{D}_{d}\in\mathbb{R}^{(D-1)\times D}$
is the first-order difference matrix along the detector direction.
This term promotes continuity between the projection-value distributions of adjacent detector elements.

The low-order DCT representation restricts the angular frequency range of dynamic SA.
We further introduce angular curvature regularization to suppress excessive oscillations within this range:
\begin{equation}
\mathcal{R}_{\mathrm{curv}}(\boldsymbol{A})
=
\frac{1}{(V-2)D}
\left\|
\boldsymbol{D}_{\theta}^{2}
\boldsymbol{\Phi}_{K}\boldsymbol{A}^{T}
\right\|_{F}^{2},
\label{eq:angular_curvature}
\end{equation}
where
$\boldsymbol{D}_{\theta}\in\mathbb{R}^{(V-1)\times V}$ and
$\boldsymbol{D}_{\theta}^{2}\in\mathbb{R}^{(V-2)\times V}$
denote the first- and second-order differences along the view sequence, respectively.
The superscript $2$ denotes the difference order, not a matrix square.
The notation $\|\cdot\|_F$ denotes the Frobenius norm.

Both true projection structures and dynamic SA may contain low angular frequencies,
so smoothness alone may still allow structural content to enter the SA branch.
To address this, as shown in the branch imposing the orthogonality constraint in Fig.~\ref{fig:framework}(C),
we first construct an orthonormal basis for the angular-gradient subspace of dynamic SA:
\begin{equation}
\boldsymbol{Q}
=
\operatorname{orth}
\left(
\boldsymbol{D}_{\theta}\boldsymbol{\Phi}_{K}
\right).
\label{eq:gradient_subspace}
\end{equation}
For the $d$th detector element,
the coordinates of their projected angular gradients in this basis are
\begin{equation}
\begin{aligned}
\boldsymbol{u}_{d}
&=
\boldsymbol{Q}^{T}\boldsymbol{D}_{\theta}
\widehat{\boldsymbol{P}}_{\boldsymbol{\Theta},:,d},
\\
\boldsymbol{v}_{d}
&=
\boldsymbol{Q}^{T}\boldsymbol{D}_{\theta}
\widehat{\boldsymbol{S}}_{\mathrm{dyn},:,d},
\end{aligned}
\label{eq:projected_gradients}
\end{equation}
where $\boldsymbol{Q}\in\mathbb{R}^{(V-1)\times K}$ and
$\boldsymbol{u}_d,\boldsymbol{v}_d\in\mathbb{R}^{K}$.
Their normalized inner product is
\begin{equation}
\rho_{d}
=
\frac{
\boldsymbol{u}_{d}^{T}\boldsymbol{v}_{d}
}{
\sqrt{
(\|\boldsymbol{u}_{d}\|_{2}^{2}+\varepsilon)
(\|\boldsymbol{v}_{d}\|_{2}^{2}+\varepsilon)
}
},
\label{eq:gradient_correlation}
\end{equation}
which defines the angular-gradient soft orthogonality constraint:
\begin{equation}
\mathcal{L}_{\perp}
=
\frac{1}{D}\sum_{d=1}^{D}\rho_{d}^{2},
\label{eq:angular_orthogonality}
\end{equation}
where $\varepsilon>0$ ensures numerical stability.
This soft constraint penalizes aligned and oppositely aligned gradients in the shared subspace, reducing ambiguity between structures and stripes.
Since the first- and second-order angular differences of fixed SA are zero,
the orthogonality and curvature constraints connected to total SA in the diagram
effectively act on the dynamic component.

As shown in Fig.~\ref{fig:framework}(C),
these constraints are introduced in two stages.
In Stage I, SA is represented only by fixed column biases,
and the IS branch parameters $\boldsymbol{\Theta}$
and fixed biases $\boldsymbol{b}$ are jointly optimized:
\begin{equation}
\mathcal{L}_{\mathrm{I}}
=
\mathcal{L}_{\mathrm{dc}}
\left(
\widehat{\boldsymbol{P}}_{\boldsymbol{\Theta}},
\boldsymbol{1}_{V}\boldsymbol{b}^{T}
\right)
+
\lambda_{\mathrm{det}}\mathcal{R}_{\mathrm{det}}
\left(
\widehat{\boldsymbol{P}}_{\boldsymbol{\Theta}}
\right).
\label{eq:stage1_loss}
\end{equation}
This stage limits the degrees of freedom of SA
and provides an initial decomposition for subsequent dynamic stripe estimation.

Stage II retains
$\boldsymbol{\Theta}$ and $\boldsymbol{b}$ from Stage I,
initializes the newly introduced dynamic coefficients $\boldsymbol{A}$ to zero,
and jointly updates all three parameter groups:
\begin{equation}
\begin{aligned}
\mathcal{L}_{\mathrm{II}}
={}&
\mathcal{L}_{\mathrm{dc}}
\left(
\widehat{\boldsymbol{P}}_{\boldsymbol{\Theta}},
\widehat{\boldsymbol{S}}
\right)
+
\lambda_{\mathrm{det}}\mathcal{R}_{\mathrm{det}}
\left(
\widehat{\boldsymbol{P}}_{\boldsymbol{\Theta}}
\right)
\\
&+
\lambda_{\mathrm{curv}}
\mathcal{R}_{\mathrm{curv}}(\boldsymbol{A})
+
\lambda_{\perp}\mathcal{L}_{\perp},
\end{aligned}
\label{eq:stage2_loss}
\end{equation}
where
$\lambda_{\mathrm{det}}$,
$\lambda_{\mathrm{curv}}$, and $\lambda_{\perp}$
are the corresponding weights.
Introducing the dynamic component and its angular constraints
after the initial separation of fixed stripes
helps prevent dynamic SA from prematurely absorbing true structures.

After optimization, the final IS and SA estimates are
$\boldsymbol{P}^{*}
=
\widehat{\boldsymbol{P}}_{\boldsymbol{\Theta}^{*}}$
and
$\boldsymbol{S}^{*}
=
\widehat{\boldsymbol{S}}
(\boldsymbol{b}^{*},\boldsymbol{A}^{*})$, respectively.
As shown in Fig.~\ref{fig:framework}(D),
$\boldsymbol{P}^{*}$
is reconstructed using filtered backprojection (FBP)
to obtain the corrected CT image.
The complete optimization procedure is summarized in Algorithm~\ref{alg:proposed_method}.

\begin{algorithm}[t]
\caption{TRACE}
\label{alg:proposed_method}
\begin{algorithmic}[1]
\REQUIRE Measured sinogram $\boldsymbol{Y}$, number of non-DC DCT basis functions $K$,
Stage I and II iteration counts $T_{1}$ and $T_{2}$,
weights $\lambda_{\mathrm{det}}$,
$\lambda_{\mathrm{curv}}$, and $\lambda_{\perp}$
\ENSURE Final IS $\boldsymbol{P}^{*}$
and SA $\boldsymbol{S}^{*}$

\STATE Initialize IS branch parameters $\boldsymbol{\Theta}$
and fixed biases $\boldsymbol{b}$

\STATE \textbf{Stage I: Fixed stripe decomposition}
\FOR{$t=1$ to $T_{1}$}
    \STATE Compute IS $\widehat{\boldsymbol{P}}_{\boldsymbol{\Theta}}$
    using~\eqref{eq:learnable_analysis_synthesis}--
    \eqref{eq:is_representation}
    \STATE Compute $\widehat{\boldsymbol{S}}_{\mathrm{fix}}$
    using the fixed component in~\eqref{eq:fixed_dynamic_sa}
    \STATE Compute $\mathcal{L}_{\mathrm{I}}$
    using~\eqref{eq:stage1_loss}
    \STATE Update $\boldsymbol{\Theta}$ and $\boldsymbol{b}$
\ENDFOR

\STATE \textbf{Stage II: Joint fixed--dynamic stripe refinement}
\STATE Retain $\boldsymbol{\Theta}$ and $\boldsymbol{b}$ from Stage I
\STATE Initialize $\boldsymbol{A}=\boldsymbol{0}$
\STATE Construct $\boldsymbol{\Phi}_{K}$ and $\boldsymbol{Q}$
using~\eqref{eq:dct_basis}
and \eqref{eq:gradient_subspace}

\FOR{$t=1$ to $T_{2}$}
    \STATE Compute IS
    $\widehat{\boldsymbol{P}}_{\boldsymbol{\Theta}}$
    \STATE Compute fixed SA, dynamic SA, and their sum $\widehat{\boldsymbol{S}}$
    using~\eqref{eq:fixed_dynamic_sa}
    \STATE Compute $\mathcal{L}_{\mathrm{II}}$
    using~\eqref{eq:stage2_loss}
    \STATE Update $\boldsymbol{\Theta}$,
    $\boldsymbol{b}$, and $\boldsymbol{A}$
\ENDFOR

\STATE Recompute the estimates using the final parameters:
$\boldsymbol{P}^{*}
\leftarrow
\widehat{\boldsymbol{P}}_{\boldsymbol{\Theta}^{*}}$
and
$\boldsymbol{S}^{*}
\leftarrow
\widehat{\boldsymbol{S}}
(\boldsymbol{b}^{*},\boldsymbol{A}^{*})$
\RETURN $\boldsymbol{P}^{*}$ and $\boldsymbol{S}^{*}$
\end{algorithmic}
\end{algorithm}

\section{Experiments}
\label{sec:experiments}

All experiments used measured PCD-CT data to retain detector-specific response errors and their angular variations. Synthetic stripes may not reproduce these patterns and can favor the assumptions of the model used to generate them. We used two complementary datasets: a QRM mouse phantom with homogeneous regions and well-defined boundaries to evaluate artifact suppression and edge sharpness, and a porcine trotter specimen with soft tissue, bone, and trabeculae to assess structure preservation in a complex background. The experiments examine the number of DCT basis functions, comparisons with representative methods, and the contribution of dynamic stripe modeling.

\begin{figure}[t]
    \centering
    \includegraphics[width=\columnwidth]{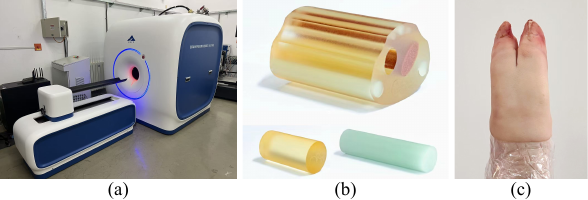}
    \caption{Experimental system and scanned objects.
    (a) In-house PCD-CT system.
    (b) QRM mouse phantom.
    (c) Porcine trotter specimen.}
    \label{fig:experimental_setup}
\end{figure}

\begin{figure*}[t]
    \centering
    \includegraphics[width=\textwidth]{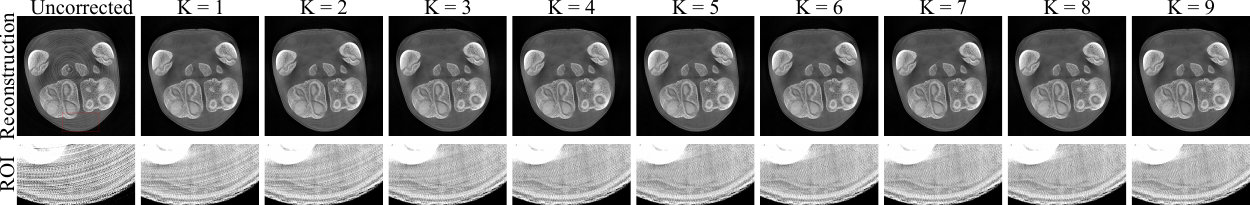}
    \caption{Effect of the number of DCT basis functions on the porcine trotter results.
    From left to right: the uncorrected result and corrected results for $K=1,\ldots,9$.
    The first row shows reconstructed images, and the second row shows magnified views of the same region marked by the red box in the uncorrected image.
    The display windows are $[0,1]$ and $[0.10,0.45]$, respectively.}
    \label{fig:dct_basis}
\end{figure*}

\begin{figure*}[t]
    \centering
    \includegraphics[width=\textwidth]{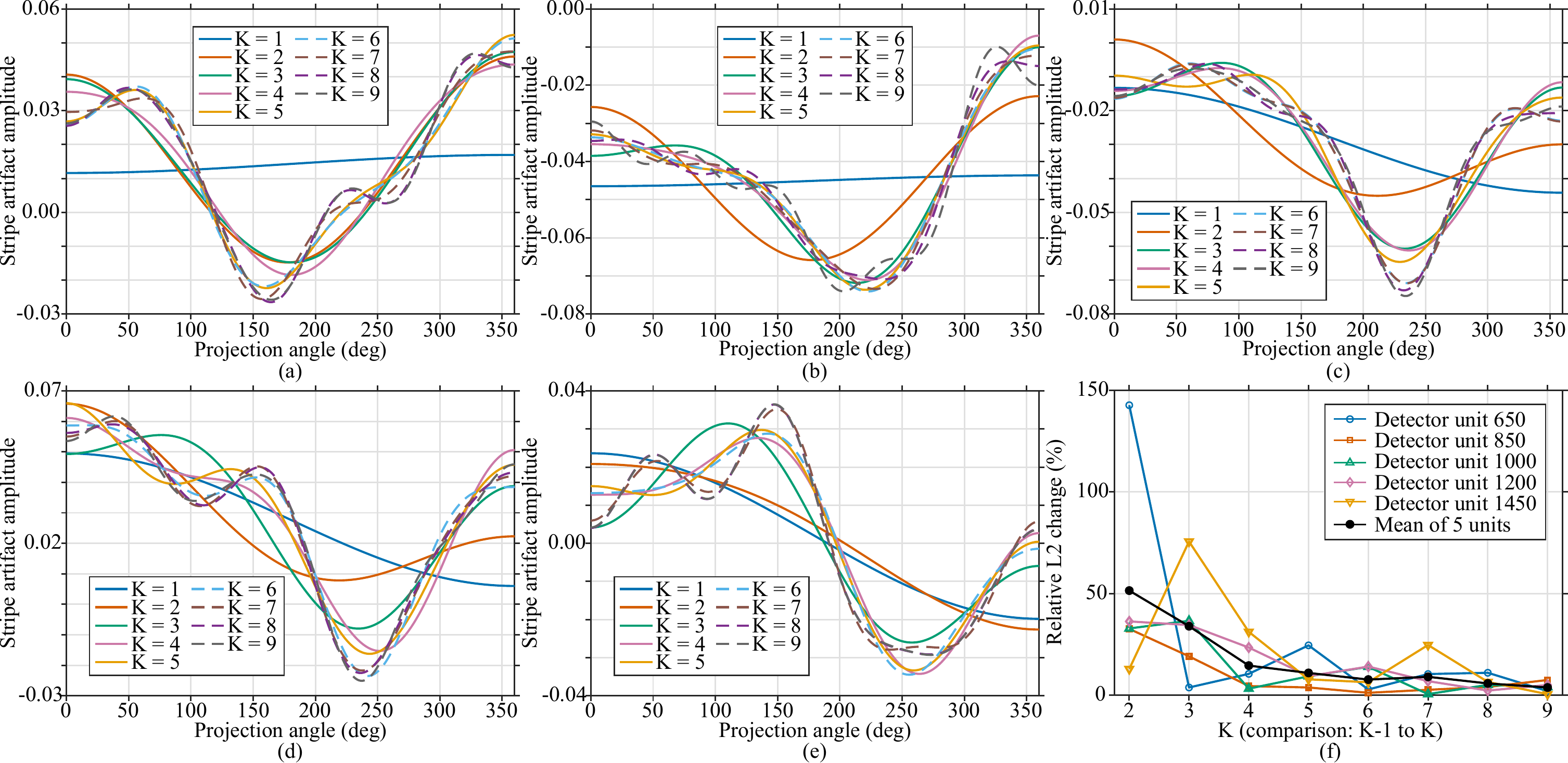}
    \caption{Effect of the number of DCT basis functions on stripe estimation.
    (a)--(e) Estimated stripe artifact amplitude versus projection angle at detector elements 650, 850, 1000, 1200, and 1450, respectively, for $K=1,\ldots,9$. The amplitude is the dimensionless logarithmic projection residual obtained by subtracting the corrected projection from the original projection.
    (f) Relative $L_2$ change (\%) between consecutive values of $K$, where the horizontal coordinate $K$ denotes the comparison from $K-1$ to $K$.
    Colored curves correspond to the five detector elements, and the black curve shows their arithmetic mean relative change.}
    \label{fig:different_K_change}
\end{figure*}

\begin{figure*}[t]
    \centering
    \includegraphics[width=\textwidth]{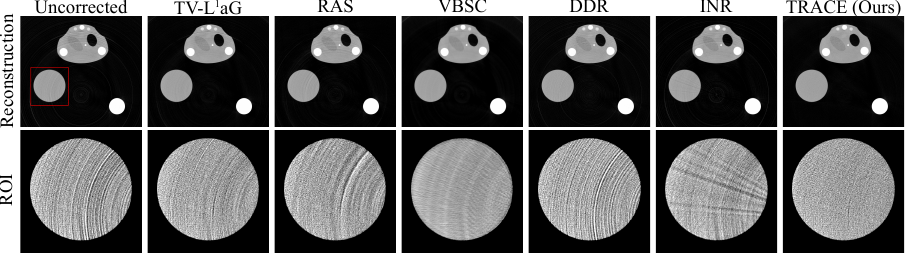}
    \caption{Ring artifact correction results of different methods on the QRM mouse phantom.
    The first row shows reconstructed images, and the second row shows magnified views of the circular water-equivalent region marked by the red box in the uncorrected image.
    The display windows are $[0,0.4]$ and $[0.2,0.28]$, respectively.}
    \label{fig:qrm_comparison}
\end{figure*}

\begin{figure*}[t]
    \centering
    \includegraphics[width=\textwidth]{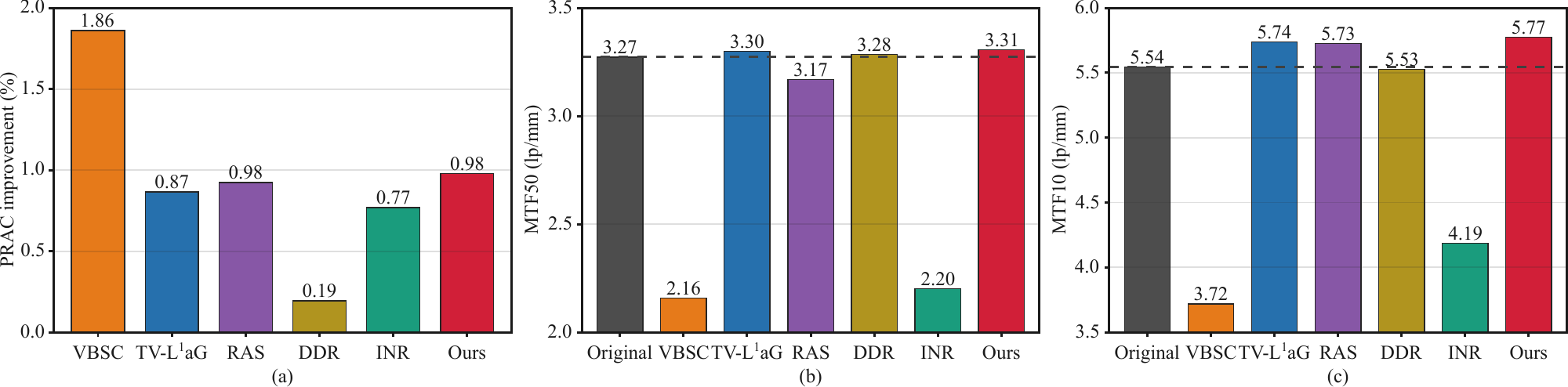}
    \caption{Quantitative evaluation of the bone-equivalent region in the QRM mouse phantom:
    (a) PRAC. (b) MTF50. (c) MTF10.
    PRAC is computed relative to the uncorrected result (Original) and reported for the five comparison methods and TRACE (labeled Ours).
    MTF50 and MTF10 are expressed in $\mathrm{lp/mm}$. Horizontal dashed lines indicate the corresponding values for Original.}
    \label{fig:qrm_results}
\end{figure*}

\begin{figure*}[t]
    \centering
    \includegraphics[width=\textwidth]{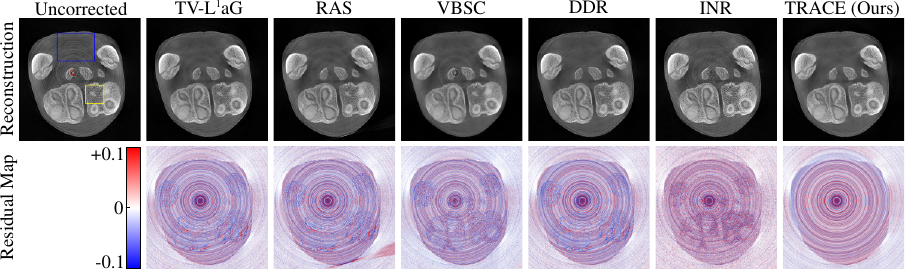}
    \caption{Ring artifact correction results of different methods on the porcine trotter specimen.
    The first row shows reconstructed images with a display window of $[0,1]$.
    Blue, yellow, and red boxes in the uncorrected image indicate regions of interest (ROIs) 1, 2, and 3, respectively.
    The second row shows image residuals, defined as the corrected reconstruction minus the uncorrected reconstruction,
    with a display window of $[-0.1,0.1]$. Red and blue indicate positive and negative values, respectively.}
    \label{fig:visual_comparison}
\end{figure*}

\begin{figure*}[t]
    \centering
    \includegraphics[width=\textwidth]{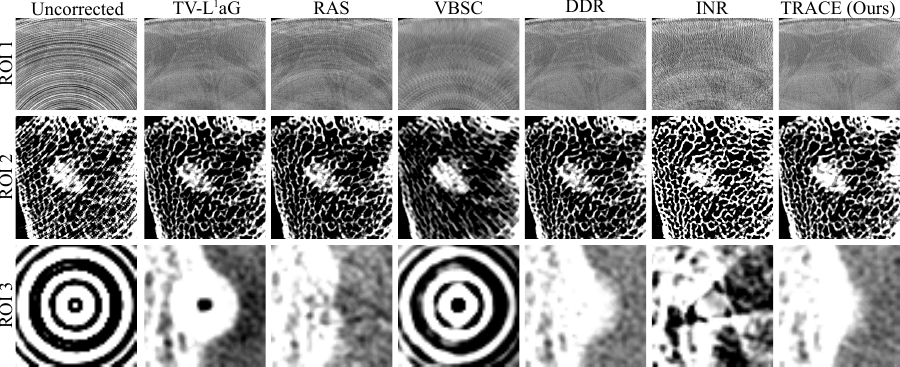}
    \caption{Magnified views of the porcine trotter specimen.
    From top to bottom: ROI~1, ROI~2, and ROI~3, corresponding to the blue, yellow, and red boxes in Fig.~\ref{fig:visual_comparison} and showing low-contrast soft tissue, trabecular bone, and the region near the rotation center, respectively.
    The display windows are $[0.1,0.5]$, $[0.4,0.6]$, and $[0.2,0.4]$, respectively.}
    \label{fig:visual_comparison_roi}
\end{figure*}

\begin{figure}[t]
    \centering
    \includegraphics[width=\columnwidth]{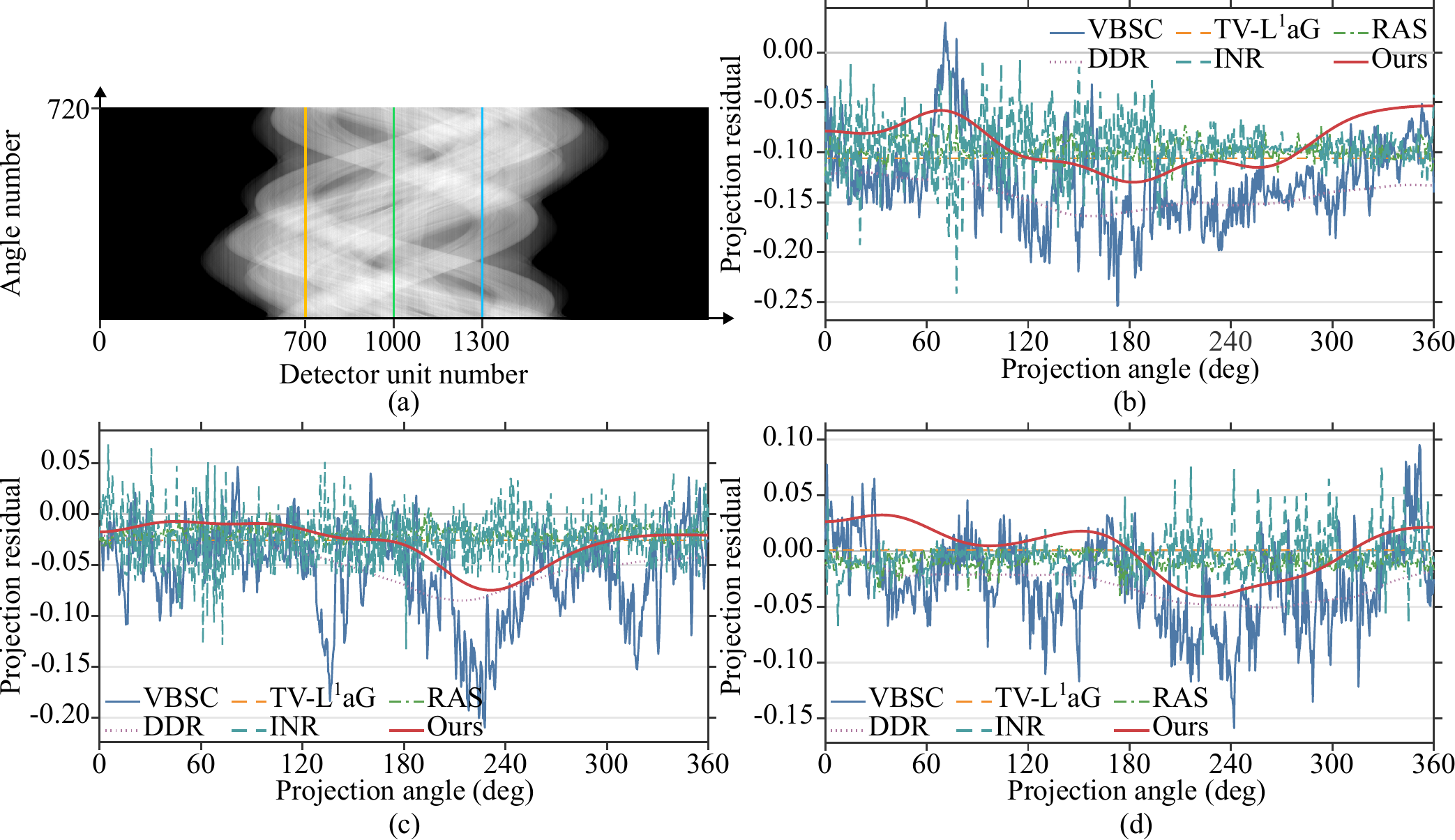}
    \caption{Angular projection residuals of different methods on the porcine trotter data.
    (a) Original sinogram, with yellow, green, and cyan vertical lines marking detector elements 700, 1000, and 1300, respectively.
    (b)--(d) Dimensionless logarithmic projection residuals versus projection angle at the corresponding detector elements.
    Residuals are defined as the original projection minus the projection corrected by each method and represent the removed projection components.}
    \label{fig:angular_residuals}
\end{figure}

\begin{figure}[t]
    \centering
    \includegraphics[width=\columnwidth]{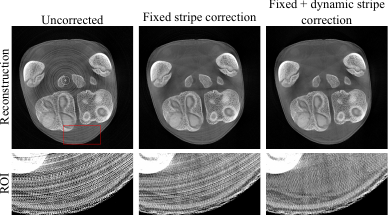}
    \caption{Effect of dynamic stripe modeling on the porcine trotter results.
    From left to right: no correction, fixed stripe correction only, and joint fixed and dynamic stripe correction.
    The first row shows reconstructed images, and the second row shows magnified views of the region marked by the red box in the uncorrected image.
    The display windows are $[0,1]$ and $[0.2,0.5]$, respectively.}
    \label{fig:ablation_dynamic}
\end{figure}

\subsection{Experimental Setup}

Data were acquired using the in-house PCD-CT system in Fig.~\ref{fig:experimental_setup}(a), comprising an L10101 X-ray source (Hamamatsu Photonics, Japan) and an EIGER2 1M-W R photon-counting detector (DECTRIS, Switzerland). The Micro-CT Mouse Phantom (QRM-70137, QRM GmbH, Germany) and porcine trotter specimen in Fig.~\ref{fig:experimental_setup}(b) and (c) were scanned at $100$ and $80~\mathrm{kVp}$, respectively. Both scans used a tube current of $300~\mu\mathrm{A}$ and an exposure time of $1~\mathrm{s}$ per view. Detector thresholds were set to $10$ and $30~\mathrm{keV}$, and only the counts recorded with the $10~\mathrm{keV}$ threshold were used.

Both scans used source-to-rotation-center and source-to-detector distances of $299.064$ and $418.657~\mathrm{mm}$, respectively, with $720$ uniformly spaced views over $360^\circ$. The central three detector rows were averaged to form fan-beam projections, and $2068$ lateral samples with a detector element width of $0.075~\mathrm{mm}$ were used for reconstruction. FBP with a Ram--Lak filter produced $2048\times2048$ images covering approximately $110.8~\mathrm{mm}\times110.8~\mathrm{mm}$, with a square pixel width of approximately $0.0541~\mathrm{mm}$.

TRACE was compared with TV-L$^{1}$aG~\cite{an2020ring}, RAS~\cite{vo2018superior}, VBSC~\cite{yan2016variation}, DDR~\cite{lu2025dual}, and INR~\cite{shi2025inr}, covering sinogram preprocessing, image postprocessing, dual-domain iteration, and unsupervised learning. Supervised methods requiring paired training data were excluded because paired artifact-free references were unavailable. TRACE is labeled Ours in the comparison figures.

QRM evaluation used the percentage of ring artifacts correction (PRAC)~\cite{zou2025prac} and the modulation transfer function (MTF). PRAC measures changes in local intensity fluctuations in polar-transformed images relative to the uncorrected result, without requiring artifact-free ground truth. Mean PRAC was computed in the homogeneous bone-equivalent region using a stabilizing constant of $C=0.5$. Higher values indicate greater artifact suppression. Since smoothing can also increase PRAC, MTF was estimated from the circular edge of the same region. MTF50 and MTF10 are the spatial frequencies at which the normalized MTF falls to $50\%$ and $10\%$, respectively, and were used to assess edge sharpness. All methods used the same rotation center, polar transformation, and evaluation region.

Porcine trotter evaluation used reconstructed images, image residuals, and magnified views of three regions of interest (ROIs): low-contrast soft tissue, trabecular bone, and the region near the rotation center. Image residuals were defined as corrected minus uncorrected reconstructions, using identical ROIs and display windows for all methods. Angular projection residuals were defined as original minus corrected logarithmic projections to show the removed projection components. TRACE used $K=7$ except in the $K=1,\ldots,9$ comparison. An ablation compared fixed stripe correction with joint fixed and dynamic stripe correction.

\subsection{Effect of the Number of DCT Basis Functions}

To assess the effect of the number of DCT basis functions, we compared the porcine trotter results for $K=1,\ldots,9$. As shown in Fig.~\ref{fig:dct_basis}, smaller values of $K$ leave visible residual rings, particularly in low-contrast soft tissue. The rings generally weaken as $K$ increases, whereas visual differences in both reconstructed images and magnified regions become small for $K\geq7$.

Fig.~\ref{fig:different_K_change}(a)--(e) shows the estimated stripe curves at five detector elements. Fewer basis functions mainly capture the overall trend. As more basis functions are included, local variations become better resolved and the main peaks and troughs stabilize. To quantify changes between consecutive settings, let $\boldsymbol{s}_{d}^{(K)}\in\mathbb{R}^{V}$ denote the original-minus-corrected projection residual at detector element $d$ using $K$ DCT basis functions. The relative change is defined as
\begin{equation}
    r_d(K)=
    \frac{\left\|\boldsymbol{s}_{d}^{(K)}
    -\boldsymbol{s}_{d}^{(K-1)}\right\|_2}
    {\left\|\boldsymbol{s}_{d}^{(K-1)}\right\|_2}
    \times100\%,\qquad K=2,\ldots,9.
    \label{eq:dct_relative_change}
\end{equation}

As shown in Fig.~\ref{fig:different_K_change}(f), the mean relative change across the five detector elements generally decreases as $K$ increases. The estimate at element 1450 still changes appreciably from $K=6$ to $K=7$, whereas the mean relative changes from $K=7$ to $K=8$ and from $K=8$ to $K=9$ are both below $10\%$. Given the limited further improvement in the reconstructed images, we selected $K=7$ to balance correction performance, stripe estimation stability, and model complexity.

\subsection{Comparison with Representative Methods}
\label{sec:comparison_representative}

We first evaluated artifact suppression and spatial resolution preservation using homogeneous regions and well-defined boundaries in the QRM mouse phantom, then assessed correction performance in the complex tissue background of the porcine trotter specimen.

\subsubsection{QRM Mouse Phantom}

Fig.~\ref{fig:qrm_comparison} shows the reconstructed QRM mouse phantom images and magnified views of the circular water-equivalent region. Ring artifacts disrupt intensity uniformity within homogeneous materials in the uncorrected image. The magnified views reveal residual arcs of varying severity with TV-L$^{1}$aG, RAS, and DDR, pronounced smoothing and blurring with VBSC, and additional linear artifacts with INR. TRACE more effectively suppresses ring-induced fluctuations in the water-equivalent region while preserving sharp boundaries.

Fig.~\ref{fig:qrm_results} presents the quantitative results for the bone-equivalent region. VBSC achieves the highest PRAC ($1.86\%$), but its MTF50 and MTF10 decrease to $2.16$ and $3.72~\mathrm{lp/mm}$, respectively. This agrees with the observed blurring and indicates that its improvement in uniformity comes at the cost of spatial resolution. TRACE achieves a PRAC of $0.98\%$, comparable to RAS and higher than TV-L$^{1}$aG, DDR, and INR. Its MTF50 and MTF10 are $3.31$ and $5.77~\mathrm{lp/mm}$, respectively, compared with $3.27$ and $5.54~\mathrm{lp/mm}$ for the uncorrected image. Together with the visual results, these findings indicate that TRACE balances ring artifact suppression and edge sharpness preservation.

\subsubsection{Porcine Trotter Specimen}

Fig.~\ref{fig:visual_comparison} shows the reconstructed porcine trotter images and image residuals. Dense ring artifacts extend across soft tissue and bone in the uncorrected image, obscuring tissue organization and fine details. The first row shows that TRACE effectively suppresses rings across different regions while preserving clear soft-tissue contours and internal bone texture. The other methods exhibit local residual artifacts, blurred details, or texture distortion.

The residual images in the second row further reveal the components altered by correction. Bone contours and tissue textures remain discernible in the residuals of some comparison methods, indicating that true structures are also affected. The residuals of TRACE are dominated by ring components with few visible traces of bone or soft-tissue structures, supporting selective artifact separation and structure preservation.

Fig.~\ref{fig:visual_comparison_roi} further compares three ROIs. In the low-contrast soft-tissue region, ROI~1, TRACE substantially reduces arc-shaped artifacts and reveals tissue organization previously obscured by rings. In the trabecular bone region, ROI~2, TRACE preserves fine trabeculae and intertrabecular spaces, whereas VBSC shows pronounced blurring and INR exhibits substantial texture distortion. Near the rotation center, ROI~3, VBSC leaves strong concentric rings, TV-L$^{1}$aG leaves a central dark spot, and INR produces pronounced local structural distortion. TRACE effectively suppresses the central artifacts while preserving local structural continuity. These results support effective ring suppression with detail preservation in complex biological tissue.

\subsection{Analysis of Angle-Dependent Stripe Artifacts}
\label{sec:angle_dependent_stripes}

Projection-domain stripe amplitudes can vary with projection angle, which a fixed column bias cannot fully describe. We examined the role of dynamic modeling using angular projection residuals and an ablation comparing fixed stripe correction with joint fixed and dynamic stripe correction.

Fig.~\ref{fig:angular_residuals} shows the projection residuals at detector elements 700, 1000, and 1300. The residuals of TV-L$^{1}$aG are nearly constant, whereas those of VBSC and INR exhibit substantial local fluctuations. The low-order DCT expansion allows TRACE to estimate a smooth, distinct angular correction profile for each detector element. For example, the estimated curve at element 1000 has a trough near $230^\circ$, whereas that at element 1300 changes from positive to negative and then rises again. These variations cannot be represented by a single fixed bias, illustrating the ability of the dynamic component to capture angle-dependent corrections.

Fig.~\ref{fig:ablation_dynamic} further demonstrates the contribution of dynamic modeling to the reconstructed images. Fixed stripe correction alone removes most strong concentric rings, but dense arc-shaped artifacts remain in the lower part of the image and are particularly evident in the magnified view. Adding the dynamic component further suppresses these artifacts and improves soft-tissue uniformity while retaining visible tissue texture and a clear outer boundary. Together, the angular residuals and reconstructed images support joint fixed and dynamic modeling for suppressing artifacts left by fixed-bias correction.

\section{Conclusion}
\label{sec:conclusion}

TRACE estimates and corrects response-related projection errors in PCD-CT through unsupervised sinogram decomposition. A low-order DCT expansion captures smooth angular stripe variations with few coefficients per detector element, while two-stage optimization and angular-gradient soft orthogonality reduce structure leakage into the artifact estimate. Experiments on measured QRM mouse phantom and porcine trotter data show that dynamic modeling suppresses rings left by fixed-bias correction. The resulting images exhibit improved uniformity with preserved edge sharpness, soft-tissue texture, and trabecular detail.

The current method uses a predefined number of DCT basis functions, and validation is limited to two-dimensional data from a single energy threshold. Future work will explore adaptive selection of the number of basis functions according to angular stripe variations and extend the framework to joint correction of multi-energy three-dimensional PCD-CT data by exploiting correlations across energy channels and detector rows.

\bibliographystyle{IEEEtran}
\bibliography{ref}

\end{document}